\documentclass[10pt,journal,compsoc]{IEEEtran}

\ifCLASSOPTIONcompsoc
  \usepackage[nocompress]{cite}
\else
  \usepackage{cite}
\fi

\usepackage{amsmath,amsfonts,amssymb}
\usepackage{algorithmic}
\usepackage{array}
\usepackage{textcomp}
\usepackage{stfloats}
\usepackage{url}
\usepackage{verbatim}
\usepackage{graphicx}

\usepackage{booktabs}
\usepackage{float}
\usepackage{multirow}
\usepackage{comment}
\allowdisplaybreaks
\DeclareMathAlphabet\mathbfcal{OMS}{cmsy}{b}{n}
\usepackage{enumerate}

\def\bA{\mathbf{A}}
\def\bB{\mathbf{B}}
\def\bC{\mathbf{C}}

\def\bff{\mathbf{f}}

\def\bI{\mathbf{I}}

\def\bK{\mathbf{K}}

\def\bX{\mathbf{X}}

\def\bZ{\mathbf{Z}}

\def\bTheta{{\boldsymbol\Theta}}

\def\bSigma{{\boldsymbol\Sigma}}

\def\bc{\mathbf{c}}

\def\bbf{\mathbf{f}}

\def\bu{\mathbf{u}}

\def\bw{\mathbf{w}}
\def\bx{\mathbf{x}}
\def\by{\mathbf{y}}
\def\bz{\mathbf{z}}

\def\balpha{{\boldsymbol\alpha}}

\def\bgamma{{\boldsymbol\gamma}}

\def\btheta{{\boldsymbol\theta}}

\def\bmu{{\boldsymbol\mu}}

\def\bxi{{\boldsymbol\xi}}

\def\bsigma{{\boldsymbol\sigma}}

\def\bpsi{{\boldsymbol\psi}}

\def\bbE{\mathbb{E}}

\def\bbR{\mathbb{R}}

\def\cL{\mathcal{L}}

\def\cN{\mathcal{N}}
\def\cO{\mathcal{O}}

\def\cS{\mathcal{S}}

\def\cV{\mathcal{V}}

\def\cov{{\sf cov}}

\def\diag{{\sf diag}}

\def\exp{{\sf exp}}

\def\KL{{\sf KL}}

\def\var{{\sf var}}

\def\zeros{\mathbf{0}}
\def\elbo{{\sf ELBO }}

\usepackage[ruled,vlined,linesnumbered]{algorithm2e} 
\usepackage{hyperref}

\usepackage{xcolor}
\hypersetup{
    colorlinks,
    linkcolor={red!50!black},
    citecolor={blue!50!black},
    urlcolor={blue!80!black}
}

\begin{document}
\title{Federated Automatic Latent Variable Selection in Multi-output Gaussian Processes}

\author{Jingyi~Gao
        and Seokhyun~Chung$^{*}$% <-this % stops a space
\IEEEcompsocitemizethanks{\IEEEcompsocthanksitem Jingyi Gao and Seokhyun Chung are with the Department of Systems and Information Engineering, University of Virginia, Charlottesville, VA USA. Corresponding author: Seokhyun Chung (E-mail: \texttt{schung@virginia.edu})\protect\\}}
% % note need leading \protect in front of \\ to get a newline within \thanks as
% % \\ is fragile and will error, could use \hfil\break instead.
% E-mail: see http://www.michaelshell.org/contact.html
% \IEEEcompsocthanksitem J. Doe and J. Doe are with Anonymous University.}% <-this % stops an unwanted space
%\thanks{*Corresponding author: \texttt{schung@virginia.edu}}}


\markboth{}
{Shell \MakeLowercase{\textit{et al.}}: Bare Demo of IEEEtran.cls for Computer Society Journals}



\IEEEtitleabstractindextext{%
\begin{abstract}
This paper explores a federated learning approach that automatically selects the number of latent processes in multi-output Gaussian processes (MGPs). The MGP has seen great success as a transfer learning tool when data is generated from multiple sources/units/entities. A common approach in MGPs to transfer knowledge across units involves gathering all data from each unit to a central server and extracting common independent latent processes to express each unit as a linear combination of the shared latent patterns. However, this approach poses key challenges in (i) determining the adequate number of latent processes and (ii) relying on centralized learning which leads to potential privacy risks and significant computational burdens on the central server. To address these issues, we propose a hierarchical model that places spike-and-slab priors on the coefficients of each latent process. These priors help automatically select only needed latent processes by shrinking the coefficients of unnecessary ones to zero. To estimate the model while avoiding the drawbacks of centralized learning, we propose a variational inference-based approach, that formulates model inference as an optimization problem compatible with federated settings. We then design a federated learning algorithm that allows units to jointly select and infer the common latent processes without sharing their data. We also discuss an efficient learning approach for a new unit within our proposed federated framework. Simulation and case studies on Li-ion battery degradation and air temperature data demonstrate the advantageous features of our proposed approach.
\end{abstract}

}

\maketitle

\IEEEdisplaynontitleabstractindextext

% \fi
\IEEEpeerreviewmaketitle

\IEEEraisesectionheading{\section{Introduction}\label{sec:introduction}}
\IEEEPARstart{T}{he} multi-output Gaussian process (MGP) is an extension of Gaussian processes (GPs) \cite{Rasmussen2006} designed to handle situations that involve data with multiple outputs. Multi-output data often arises in the Internet of Things (IoT)-enabled connected systems \cite{Kontar2021}, where several units or devices collect data individually from their operation. Beyond modeling correlations within a single unit, the covariance structure in MGPs can capture correlations across different units based on their data. This capability allows for extracting relatedness between units as well as identifying unit-specific features. Within this framework, knowledge transfer across different units becomes feasible, often leading to enhanced predictive performance compared to independent modeling of individual units. This advantageous feature further comes with non-parametricity and the ability to quantify predictive uncertainty inherited from the natural Bayesian interpretation of GPs. As a result, MGPs have been broadly used across various domains. Examples include forecasting the energy demand of EV charging stations \cite{gilanifar2020}, providing individualized failure event predictions for forklift trucks in a telematics system to enable timely part replacements \cite{jahani2021}, and offering situational awareness support for disaster response using a network of weather sensors \cite{osborne2008}.

One popular strategy for building an MGP is to introduce a set of independent latent variables shared across units. The key intuition is to use the latent variables to model common latent functions across multiple units. Methods along this line include the linear model of coregionalization (LMC) \cite{Alvarez2012} or convolution process (CP)-based covariance construction \cite{Alvarez2008}. LMC-based methods express outputs as a linear combination of the latent functions, while CP-based methods model outputs as a sum of CPs that convolve the shared latent functions on kernels specific to each output. By doing so, these approaches can build a valid covariance matrix that estimates dependencies between outputs. 

However, the increased connectivity through recent IoT technologies has realized the deployment of large-scale connected systems that involve many units potentially operated under diverse environments or configurations \cite{Ray2017}, which induces a key challenge in extracting commonality across units using latent variables. For instance, connected car systems involve vehicles of different types (e.g., trucks or sedans) that navigate through different driving conditions (e.g., in-town or highway). In such situations, dependencies between units become complex. An integrative analysis thus demands an understanding of non-trivial dependencies among units inherent in their data, which a few latent functions cannot readily capture. Merely increasing the number of shared latent functions may not be a desired solution, as it leads to a high-dimensional, non-convex search space in model inference. An MGP with many latent functions is susceptible to being trapped in poor local optima, slowing down convergence, or overfitting when training observations are scarce. This trade-off between generalizability and complexity explains why most literature constrains the number of latent functions to be small (one to four) \cite{Moyan2022}, and finding the adequate number still resorts to a brute-force approach. 

Another critical challenge in MGP-based analysis for IoT-enabled connected systems is attributed to its typical practices that assume a centralized framework \cite{Li2020}. Commonly, connected systems are configured in a hub-and-spoke architecture, where a central server located on a cloud platform wirelessly communicates with peripheral edge units. In this setup, centralized data analysis requires the aggregation of data from all edge units into a centralized database which is then transmitted to the central server for processing and model training using MGPs. However, this centralized approach likely encounters several challenges, including scalability constraints, data privacy concerns, and potential bottlenecks in data transmission. Specifically, significant communication demands arise when multiple units transmit their data to the central server, and substantial computing resources are required for centralized data processing. Furthermore, the privacy of individual units is compromised as they share their data with the central server. These challenges are compounded by the inherent inscalability of MGPs, characterized by cubic complexity with respect to the number of data points across all units.

In this article, we aim to address the challenges above arising in MGPs: (i) difficulty in determining an adequate number of latent functions, and (ii) issues attributed to the traditional centralized regime. To this end, we develop a framework to build an MGP that can automatically select a subset of shared latent functions essential for representing inherent patterns common across units. Furthermore, our approach allows for its inference and prediction through federated learning (FL), leveraging distributed computing resources across units without sharing raw data.

Specifically, we build an LMC-based hierarchical model tailored for the hub-and-spoke structure of connected systems, where the common latent functions are shared via a central server, while unit-specific coefficients are situated in the respective units. For each unit, we place spike-and-slab priors on its latent function coefficients. The intuition is that the shrinkage prior imposes sparsity over the coefficients, which only maintain a necessary number of latent functions by shrinking the coefficients for other insignificant ones to zero. 

For model estimation, we propose an inference method that builds upon variational Bayesian approximation \cite{Titsias2009, Matthew2013}. Our approach derives a lower bound for the marginal loglikelihood. Maximizing this lower bound in turn approximates the intractable original posterior as well as estimates model parameters. In particular, one of our key contributions lies in the derivation of a lower bound that can be maximized by the FL framework. This lower bound allows us to develop an FL-based inferential algorithm, which harnesses the computing power of individual units without sharing raw data or unit-specific information at any stage. We further show that our framework can seamlessly integrate a simple yet efficient learning method for units newly entering the system after model training is completed, without using the computational resources of other units.

Overall, the contributions of this study can be summarized as follows:
\begin{itemize}
\item {We build an LMC model that (i) does not require raw data sharing in estimating cross-unit covariances as well as (ii) can automatically infer the necessary number of latent functions to represent the shared pattern across units. These objectives are achieved by leveraging the hub-and-spoke structure, an inducing point approach, and a sparse prior placed on the coefficients of the latent functions.}
\item {We propose an FL framework that estimates the parameters of our proposed model and makes a personalized prediction for each unit, without requiring centralized computation and disclosing units' local data. Our framework exploits units' local computing power, thereby addressing various fundamental issues arising from traditional centralized learning frameworks for MGPs.}
\item {We propose a learning approach for new units entering the system, which allows them to leverage information from other units without disclosing their own data. This approach synergies with our proposed automatic latent function selection by focusing only on the selected latent functions when learning for the new unit.}
\item {We show the effectiveness of our proposed model in real-world applications using reliability engineering and climate data.}
\end{itemize} 

The rest of the paper is organized as follows. In Section \ref{s:related}, we review the related literature. In Section \ref{s:lmc}, we discuss preliminaries and challenges to be addressed. In Section \ref{s:proposed}, we build our model and present its inference and prediction. In Section \ref{s:experiment}, we discuss numerical studies using simulation and real-world data. In Section \ref{s:discussion_conclusion}, we draw a conclusion.

\section{Related Work} \label{s:related}
\subsection{Literature on MGPs}
The MGP was originally known as \textit{cokriging} in the geostatistics field \cite{Kleijnen2009}, where it gained significant attention as an effective tool for modeling the relatedness between regionalized variables. Beyond geostatistics, the ability to handle multivariate data has allowed MGPs to achieve notable success in broader fields, including healthcare (e.g., \cite{Cheng2020}), transportation (e.g., \cite{Rodrigues2018}), system prognostics (e.g., \cite{Kontar2017}), and ecology (e.g., \cite{Ingram2020}).  This broad applicability has spurred research into designing MGPs to address specific challenges in different scenarios. For instance,  Moreno-Munoz et al. \cite{Moreno2018} developed a heterogeneous MGP, designed for modeling outputs with heterogeneous data types, such as categorical and time series outputs. Wang et al. \cite{Wang2022} developed an MGP that can handle input domain inconsistency and negative knowledge transfer \cite{Kontar2020} occurring due to a lack of shared information among some outputs. Chung et al. \cite{Chung2022} proposed a weakly-supervised MGP to address cases where output membership is unknown for some observations. Additionally, Soleimani et al. \cite{soleimani2017scalable} introduced an approach that combines an MGP and a survival analysis model for the joint modeling of longitudinal and time-to-event data.

While there are many success stories, a prominent challenge in MGPs is their instability with datasets containing numerous observations. This issue stems from standard GPs that suffer $\cO(N^3)$ time complexity when modeling $N$ observations due to the inversion of an $N\times N$ covariance matrix. This complexity is extended to $\cO(M^3N^3)$ for MGPs when modeling $M$ outputs, each with $N$ observations. To tackle this issue, several approaches have been proposed. For example, inspired by sparse approximation of GPs \cite{Snelson2005}, Alvarez and Lawrence \cite{Alvarez2008} introduced a sparse approximation for the shared latent processes using pseudo-inputs, which reduces the complexity of inverting the approximated covariance matrix. This approach was later extended by Alvarez et al. \cite{Alvarez2010}, who proposed a variational inference (VI)-based framework to approximate the posterior of the sparse latent processes, addressing the limitations of assuming smooth shared latent functions. Nguyen
et al. \cite{Nguyen2014} developed a scalable MGP model built upon the LMC with sparse approximation, amenable to stochastic VI \cite{hoffman2013stochastic} that facilitates the use of mini-batches in optimization. Recently, Bruinsma et al. \cite{Bruinsma2020} showed that the orthogonal instantaneous linear mixing models can achieve scalable inference and learning even without sparse approximation, offering an alternative approach to address scalability issues.

As demonstrated by the studies mentioned above and others (e.g., \cite{Dahl2019, Requeima2019, Zhe2019}), modeling each output as a linear combination of shared latent functions is highly popular in constructing an MGP. However, the literature on determining the appropriate number of latent functions for specific data is quite scarce. Notably, Titsias and Lazaro-Gredilla \cite{Titsias2011} introduced a method for selecting latent functions using spike-and-slab priors. However, this method is only applicable within a centralized regime and scales poorly with the number of observations. While our model also automatically selects latent functions, its unique advantages lie in its compatibility with federated systems and scalability through stochastic optimization.

\subsection{Literature on GPs in federated settings}
With recent revolutionary advances in technologies that miniaturize chips with computing power, units at the edge, which used to merely create or collect data, increasingly possess significant computing and data storage capabilities. FL, since its first introduction by McMahan et al. \cite{mcmahan2017}, has garnered explosive interest in recent years as a collaborative framework to train models using the local computing resources of multiple units while keeping their data stored locally. This approach effectively addresses various challenges in FL such as data/system heterogeneity \cite{Ye2023}, client drifts \cite{karimireddy2020, Koyejo2022}, fairness \cite{huang2024}, and more. Despite these advancements, the majority of FL literature has focused predominantly on deep learning models \cite{hao2019}. 

Recently, several studies have aimed to extend FL methods beyond deep learning. A line of these efforts focused on GPs in federated settings. Achituve et al. \cite{Achituve2021} studied personalized FL by collaboratively estimating a GP with deep kernels across units and employing personalized GP classifiers for each unit. Yu et al.  \cite{Yu2022} investigated deep kernel learning in federated settings but avoided kernel sharing across units by working directly in the feature space. Yue and Kontar \cite{Yue2024} explored a method that directly estimates GP hyperparameters via FL and studied its theoretical properties. Furthermore, Zhang et al. \cite{Zhang2022} examined GP regression under federated settings in the presence of units subject to Byzantine attacks, while Guo et al. \cite{guo2022} investigated a framework that learns a sparse GP model using FL algorithms. Note that these studies primarily focused on learning a single-output GP under federated settings. 

In contexts where units in a federated regime have separate but correlated GPs, our study relates to recent work by Chung and Kontar \cite{Chung2023}. Both studies establish an FL framework for MGPs where each output corresponds to a different unit. However, our model offers distinctive advantages: it can infer the appropriate number of latent variables needed to extract common latent functions across units, and thus, enhance prediction accuracy for each unit as well as enable efficient prediction for a new unit by using latent functions selectively instead of using all of them.

\section{Brief Review of LMC}\label{s:lmc} 
Consider a connected system with $M$ units indexed by $m \in \{1,...,M\}$, where unit $m$ collects data $\mathbfcal{D}_m= (\bX_m, \by_m)$ with inputs $\bX_m= [\bx_{m,n}]_{n=1,...,N_m}^\top \in \bbR^{N_m\times d}$ and outputs $\by_m = [y_{m,n}]^\top_{n=1,...,N_m} \in \bbR^{N_m \times 1}$. Collectively, we define $\mathbfcal{D} = \{\mathbfcal{D}_m\}_{m=1}^M$ and $\by = [\by_m^\top]^\top_{m=1,...,M}$. Note that, for notation brevity, we temporarily assume that the units have the same number of observations $N=N_1=\cdots=N_M$, yet this assumption will be relaxed in subsequent sections. 

Given this setting, a multi-output regression problem estimates $M$ different underlying regression functions $\{f_m\}_{m=1}^{M}$,  where $f_m:\bbR^d \rightarrow \bbR$ relates the inputs $\bX_m$ to outputs $\by_m$, while considering dependencies between functions. Specifically,
\begin{equation}\label{eq:out_func}
y_{m,n} = f_m (\bx_{m,n}) + \epsilon_{m} %\; \text{ for } \; n\in \{1,...,N\},    
\end{equation}
where $\epsilon_{m} \sim \cN(0, \sigma_m^2)$ is a random Gaussian noise. We will call the unknown regression function $f_m$ as ``output function''. To model dependencies between outputs, an MGP builds an extended GP that assumes the output function values across all units are jointly Gaussian distributed. Denoting the output function values by $\bbf = [\bbf_m^\top]^\top_{m=1,...,M}$ with 
$\bbf_m = [f_m( \bx_{m,n})]^\top_{n=1,...,N} = [f_{m,n}]^\top_{n=1,...,N}$, an MGP can be expressed as a multivariate Gaussian:
\begin{align}
p(\bff\vert \bX) &= \cN (\bff;\zeros, \bC_{\bff,\bff}) \nonumber \\ &= \cN \left(\bff;\zeros,  \begin{bmatrix}
\bC_{\bff_1, \bff_1}& \cdots & \bC_{\bff_1, \bff_M} \\
\vdots & \ddots& \vdots \\
\bC_{\bff_M, \bff_1} & \cdots & \bC_{\bff_M, \bff_M}
\end{bmatrix} \right)    \label{eq:mgp}
\end{align}
where we use the notation $\bC_{\cdot, \cdot}$ to denote a (cross-) covariance matrix for the corresponding variables that appear at the subscript, which will be used throughout the paper. For example, $\bC_{\bff_m,\bff_{m'}}$ is the covariance matrix between the output $m$ and $m'$, where its elements are calculated by $\cov[f_m(\bx_{m,n}), f_{m'}(\bx_{m',n'})] = k_{m, m'}(\bx_{m,n}, \bx_{m',n'};\bxi_{m,m'})$ for $n,n' \in \{1,...,N\}$ with a kernel $k_{m, m'}(\cdot, \cdot; \bxi_{m,m'} )$ parameterized by $\bxi_{m,m'}$. 

Establishing a valid MGP in \eqref{eq:mgp} necessitates ensuring $\bC_{\bff,\bff}$ to be positive semi-definite. The LMC provides an intuitive way to understand between-unit dependencies with a guarantee of the positive semi-definitiveness of $\bC_{\bff,\bff}$. The idea is to introduce a set of ``latent functions'' shared across units $m \in \{1,\dots,M\}$. Specifically, the LMC expresses an output function as a linear combination of $L$ common latent functions, represented as
\begin{align}\label{eq:linear_comb}
f_m(\bx_{m,n}) = \sum_{l=1}^L w_{m,l} u_l(\bx_{m,n}) \; \text{ for } n \in \{1,...,N\}\;
\end{align}
where $u_l(\cdot)$ represents the $l$-th latent function for $l \in\{1,...,L\}$ and $w_{m,l}$ is the scalar coefficient associated with both $u_l(\cdot)$ and $f_m(\cdot)$. Here the shared latent functions are modeled as respective GPs independent of each other. Each $u_l(\cdot)$ is characterized by mean zero and covariance $\cov(u_l(\bx), u_{l}(\bx')) = k_l(\bx, \bx'; \bxi_l)$ with the covariance function $k_l(\cdot, \cdot; \bxi_l)$ parametrized by $\bxi_l$, while $\cov(u_l(\bx), u_{l'}(\bx')) = 0$ for $l\neq l'$. Given this modeling approach, the (cross-) covariance $\bC_{\bff_m, \bff_{m'}} \in \bbR^{N\times N}$ has the $(n, n')$-th element calculated by
\begin{align}\label{eq:lmc_cov}
    &\cov\left[f_m(\bx_{m,n}), f_{m'}(\bx_{m',n'})\right] \nonumber\\
    &= \sum_{l=1}^{L}\sum_{l'=1}^L w_{m,l}w_{m',l'} \cov\left[u_l(\bx_{m,n}), u_{l'}(\bx_{m'n'})\right] \nonumber\\ 
    &= \sum_{l=1}^{L} b_{m,m'}^l k_l(\bx_{m,n}, \bx_{m',n'}; \bxi_l)
\end{align}
where the second identity is due to the independence between latent functions $u_l(\cdot)$ and $u_{l'}(\cdot)$ for $l\neq l'$, and $b_{m,m'}^l =  w_{m,l}w_{m',l}$.  Given \eqref{eq:lmc_cov} we can write the (cross-) covariance $\bC_{\bff_m, \bff_{m'}} = b_{m,m'}^l\bC_{\bu_{m,l}, \bu_{m',l'}}$ where $\bu_{m,l} = [u_l(\bx_{m,n})]^\top_{n=1,...,N}$ indicates the vector of latent function values evaluated at $\bX_m$. Here a positive semi-definite matrix $\bB_l \in \bbR^{M\times M}$ with the $(m, m')$-th element $b_{m,m'}^l$ is often coined as the coregionalization matrix. We will let $\bC^{\sf LMC}_{\bff,\bff}$ denote the MGP covariance matrix $\bC_{\bff,\bff}$ in \eqref{eq:mgp} constructed using the covariance function in \eqref{eq:lmc_cov}. 

In the literature on LMCs, one common approach for model estimation is the maximum likelihood method. It seeks to solve an optimization problem: $\max_{\bTheta} \log p(\by\vert \bX; \bTheta)$, which optimizes $\bTheta$ that indicates all hyperparameters involved. Given \eqref{eq:out_func} and \eqref{eq:mgp}, the log-likelihood of the LMC model can be written as 
\begin{align}
    &\log p(\by\vert \bX; \bTheta) = \log \int p(\by\vert \bff)p(\bff \vert \bX)d\bff \nonumber\\
    &\> = \log \cN\left(\by;\zeros, \bC_{\bff,\bff}^{\sf LMC} + \bSigma\right) \nonumber\\ 
    &\> = -\frac{1}{2} \by^\top (\bC_{\bff,\bff}^{\sf LMC} + \bSigma)^{-1} \by -\frac{1}{2} \log \vert  \bC_{\bff,\bff}^{\sf LMC} + \bSigma \vert - const \nonumber
\end{align}  
where $\bSigma = \diag(\bsigma) \otimes \bI_N$ with $\diag(\bsigma)$ being the diagonal matrix corresponding to $\bsigma = [\sigma_m]_{m=1,...,M}^\top$. In this case, $\bTheta = \{\{\bB_l\}_{l=1}^L, \{\bxi_l\}_{l=1}^L, \bsigma\}$.

For a new input $\bx_* \in \bbR^d$, the predictive distribution of $y_{m,*}$ for output $m$ is derived by
\begin{align}\label{eq:pred}
       &p(y_{m,*} |\bx_*, \bX, \by) = \int p(y_{m,*} |\bff, \bx_*, \bX, \by) p(\bff |\bX, \by) d\bff \nonumber\\ 
       & = \cN\left(y_{m,*}; \bc_*(\bC^{\sf LMC}_{\bff,\bff}+\bSigma)^{-1}\by, \right. \nonumber\\
       &\quad\quad\quad\quad\quad\quad\quad
       \left.c_{*,*} - \bc^\top_*(\bC^{\sf LMC}_{\bff,\bff}+\bSigma)^{-1} \bc_* + \sigma_m^2 \right)
\end{align}
with $c_{*,*} = \cov\left[f_m(\bx_*), f_{m}(\bx_*)\right]$ and $\bc_{*} = [\bc^\top_{*,m'}]^\top_{m'=1,...,M} \in \bbR^{NM \times 1}$ where $\bc_{*,m'}\in \bbR^{N \times 1}$ is the vector collects $N$ covariances between $f_m(\bx_*)$ and $f_{m'}(\bx_{m',n})$ for $n=1,...,N$, i.e., $\bc_{*,m'}=\left[\cov\left[f_m(\bx_*), f_{m'}(\bx_{m'n})\right]\right]^\top_{n=1,...,N}$. 

\subsection{Challenges}\label{s:challenges}
In contrast to independent modeling for individual units, the LMC is advantageous in analyzing connected systems that involve multiple units by accounting for dependencies across units. However, several challenges are encountered when it is directly deployed to connected systems in practice. 

\textbf{Determining $L$}. Identifying non-trivial commonalities across units demands a sufficient number of latent functions to help construct a rich correlation structure capable of estimating complex dependencies between outputs. Yet, having more latent functions can significantly increase the risk of overfitting and computational burden. Thus, determining an adequate $L$ becomes crucial. A naive brute-force search for $L$ is substantially inefficient, given the computational complexity $\cO(N^3M^3)$ of MGPs.

\textbf{Centralized model building and learning.} The LMC model construction and inference implicitly assume a centralized process for managing all data and model parameters. That is, model building and training are done at a central location (e.g., a central server on a cloud) that amasses data from all units. To see this, building the covariance matrix $\bC^{\sf LMC}_{\bff,\bff}$ needs to calculate $\sum_{l=1}^{L} b_{m,m'}^l k_l(\bx_{m,n}, \bx_{m',n'}; \bxi_l)$ in \eqref{eq:lmc_cov}, which requires sharing observations $\bx_{m,n}, \bx_{m',n'}$ as well as the model parameter $b_{m,m'}^l$. Furthermore, \eqref{eq:pred} shows that the derivation of the predictive distribution needs a posterior distribution $p(\bff \vert \bX, \by) = p(\bff \vert \mathbfcal{D})$ conditioning on data across all units $\mathbfcal{D}$, again implying the need for a centralized process. Unfortunately, such centralized frameworks have critical drawbacks attributed to (i) the communication inefficiency induced by transmitting units' raw data to the central server, (ii) the compromised privacy due to data sharing, and (iii) the need for massive computational power at the central cloud, often overwhelmed in a large IoT-enabled connected system that involves a lot of units.

\section{Proposed Approach} \label{s:proposed}
This section introduces our proposed approach to building an MGP framework that addresses the challenges above. Section \ref{s:hierarchical} discusses our hierarchical Bayes-based approach that builds an LMC upon the hub-and-spoke structure, equipped with priors that enable automatic latent function selection. Section \ref{s:federatedvi} discusses our framework for federated model inference, which exploits local computing resources of units without sharing their raw data and the need for centralized computation. Section \ref{s:new_units} discusses how our distributed framework enables an efficient learning method for a unit that newly joins the system.

\subsection{Hierarchical MGP with Spike-and-slab Priors}\label{s:hierarchical}
One important characteristic of the LMC is that it features a hierarchical structure, where $f_m$ can be viewed as generating from the shared latent functions $\{u_m\}_{l=1}^L$ along with unit-specific parameters $\{w_{m,l}\}_{l=1}^L$. This implies that $\{f_m\}_{m=1}^M$ are conditionally independent of each other when the entire length of $\{u_l\}_{l=1}^L$ is given. Such hierarchical structure aligns with the hub-and-spoke structure of IoT-enabled connected systems, where multiple units at the edge are connected to the common central server. Given this structural accordance, we build an LMC-based hierarchical Bayes model that mirrors the same hub-and-spoke structure with connected systems, while incorporating a prior distribution that imposes sparsity on the coefficient parameters $\{w_{m,l}\}$. The prior distribution encourages some coefficients to be shrunk to zero so that only a set of latent functions needed to capture between-unit dependencies remains active. Interestingly, our hierarchical modeling allows for getting around the direct calculation of \eqref{eq:lmc_cov}. This, in turn, facilitates to build an FA-based inferential algorithm (discussed in Section \ref{s:federatedvi}) where units collaboratively train the hierarchical model while sharing only needed information with the central server and securing unit-specific information that may compromise privacy if disclosed.

Our hierarchical modeling starts with placing a prior, which imposes sparsity on the coefficients $\bw=[\bw_m^\top]^\top_{m=1,\cdots,M}$ with $\bw_m=[w_{m,l}]_{l=1,\cdots,L}^\top$ in \eqref{eq:linear_comb}. Specifically, we place a spike-and-slab prior, written as 
\begin{align}
    p(\bw)&=\prod_{m=1}^M\prod_{l=1}^L p(w_{m,l}) \nonumber\\
    &= \prod_{m=1}^M\prod_{l=1}^L \pi\cN\left(0, \sigma^2_{\bw}\right)+(1-\pi)\delta_0(w_{m,l}) \label{eq:ssp}
\end{align}
where identical priors are independently placed on $w_{m,l}$. The spike-and-slab prior is characterized by a mixture of the Dirac delta $\delta_0(w_{m,l})$ centered at zero and a Gaussian distribution with a mean of zero and a variance of $\sigma^2_{\bw}$, with a mixing probability $\pi$. Intuitively, the prior indicates that $w_{m,l}$ follows a Gaussian distribution centered at zero with probability $\pi$, or is simply zero with probability $1-\pi$. In the presence of $\delta_0(w_{m,l})$, the prior puts a substantial mass on zero. This results in encouraging some $w_{m,l}$ to be zero in model inference, thereby eliminating the influence of $u_l$ in characterizing $f_m$. Consequently, a set of only necessary $u_l$ in describing relatedness across output functions $\{f_m\}_{m=1}^M$ can be identified.

Now, let us tackle the issue of the direct calculation of covariances in \eqref{eq:lmc_cov}. We want to estimate the covariance without the need to share inputs across units. Our idea is based on the sparse MGP method \cite{Alvarez2008}. Specifically, it introduces a set of \textit{inducing} or \textit{auxiliary variables}, denoted by $\bZ_l = [\bz_{l,q}]^\top_{q=1,...,Q} \in \bbR^{Q\times d}$, at which the latent functions are evaluated, i.e., $\tilde \bu_l = [u(\bz_{l,q})]^\top_{q=1,...,Q}$. Here $\{f_m(\cdot)\}_{m=1}^M$ are assumed to be conditionally independent given $\tilde \bu_l$, in place of the entire information of $\{u_l(\cdot)\}_{l=1}^L$. The rationale is that $\tilde \bu_l$ may characterize $u_l(\cdot)$ sufficiently well. This method indeed originates from the effort to reduce the computational complexity of MGPs by choosing $Q \ll N$ \cite{Snelson2005}. Interestingly, it in turn eliminates the direct calculation of covariance in the LMC. To see this, we present 
\begin{align}
p(\tilde{\bu}) &= \cN(\tilde{\bu};\zeros,\bC_{\tilde{\bu},\tilde{\bu}})=\prod_{l=1}^L\cN\left(\tilde{\bu}_l;\zeros, \bC_{\tilde{\bu}_l,\tilde{\bu}_l}\right) \label{eq:p(tildeu)}\\
p(\bbf\vert\bw,\tilde{\bu})&= \prod_{m=1}^{M}p(\bbf_m\vert\bw_m,\tilde{\bu}) \nonumber\\
&= \prod_{m=1}^{M}\cN(\bff_m;\bC_{{\bff_m, \tilde{\bu}}}\bC_{\tilde{\bu}, \tilde{\bu}}^{-1}\tilde{\bu}, \nonumber\\
&\quad\quad\quad\quad\quad\quad\quad
\bC_{\bbf_m, \bbf_m}-\bC_{{\bff_m, \tilde{\bu}}}\bC_{\tilde{\bu}, \tilde{\bu}}^{-1}\bC_{\tilde{\bu}, \bff_m}) \nonumber\\
&= \prod_{m=1}^{M}\cN\left(\bff_m;\sum_{l=1}^Lw_{m,l} \bA_{m,l}\tilde{\bu}_l, \sum_{l=1}^Lw^2_{m,l}\bK_{m,l}\right)
\label{eq:p(f|w,g)}
\end{align} 
with 
\begin{align*}
\bA_{m,l} &=  \bC_{{\bu_{m,l}, \tilde{\bu}_l}}\bC_{\tilde{\bu}_l, \tilde{\bu}_l}^{-1}, \\ \bK_{m,l} &= \bC_{\bu_{m,l}, \bu_{m,l}}-\bA_{m,l}\bC_{\tilde{\bu}_l, \bu_{m,l}}
\end{align*}
where \eqref{eq:p(tildeu)} represents the latent independent GPs for $\tilde\bu_l$ evaluated at $\bZ_l$ and \eqref{eq:p(f|w,g)} implies the conditional independence. It is important to note that covariance matrices directly calculating covariances between units (e.g., $\bC_{\bff_m, \bff_{m'}}$ or $\bC_{\bu_{m,l}, \bu_{m',l}}$ for $m \neq m'$) no longer appear in \eqref{eq:p(f|w,g)} and \eqref{eq:p(tildeu)}. As such, we can calculate \eqref{eq:p(f|w,g)} and \eqref{eq:p(tildeu)} without sharing data across units. 

Finally, the probability distribution of outputs can be represented as 
\begin{equation} \label{eq:p(y|f)}
p(\by\vert\bbf)=\prod_{m=1}^{M}\cN(\by_m;\bbf_m,\sigma_m^2\bI)
\end{equation}
given regression modeling with the Gaussian noise in \eqref{eq:out_func}. 

\begin{figure}[htb!]
    \centering    
    \includegraphics[width=\columnwidth]{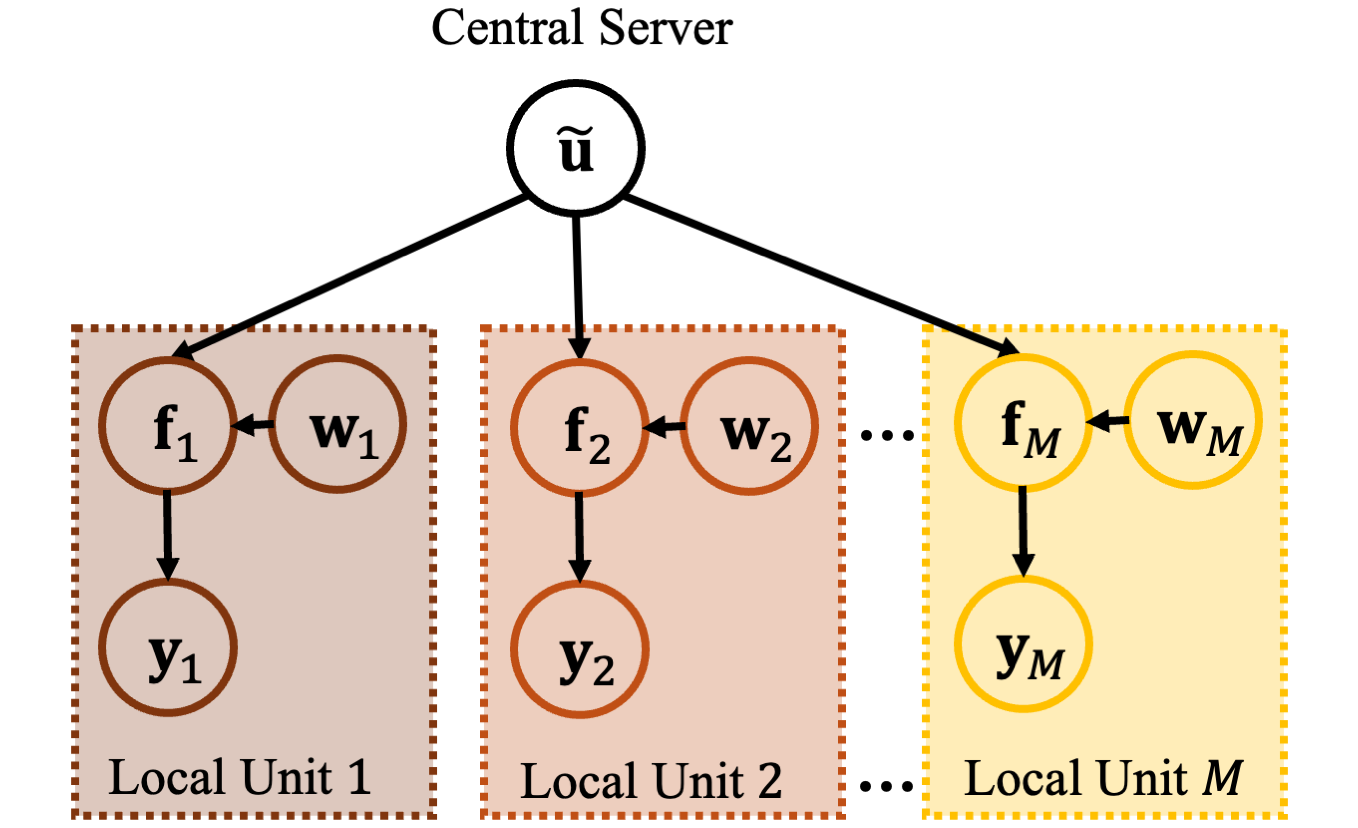}
    \caption{Our hierarchical modeling for LMC with spike-and-slab priors and its structural correspondence to connected systems.}
    \label{fig:hier}
\end{figure}

The complete form of our hierarchical model is given by \eqref{eq:ssp}, \eqref{eq:p(tildeu)}, \eqref{eq:p(f|w,g)} and \eqref{eq:p(y|f)}. Figure \ref{fig:hier} illustrates the hierarchical structure of our model. One should note that the structure aligns with the natural hierarchy of IoT-enabled connected systems. Based on this structure, our proposed model can estimate the relatedness across units while situating their data $\mathbfcal{D}_m$ and unit-specific parameters $\bw_m$ in the storage of respective units at the edge. Instead, only needed information ($\tilde\bu_l$ and $\bZ_l$) is shared through the central server. 

Unfortunately, despite the hierarchical model construction, the inference and prediction are still troublesome because (i) predictions using the proposed model involve a posterior distribution given \textit{all} data from units, which requires the collection of local datasets to a central location and (ii) the posterior distribution is intractable. These challenges can be clearly seen in the posterior distribution of our model, represented as
\begin{equation}
p(\bbf,\tilde{\bu},\bw\vert\by, \bX,\bZ) = \frac{p(\by\vert\bbf,\tilde{\bu},\bw,\bX,\bZ)p(\bbf,\tilde{\bu},\bw)}{p(\by\vert\bX,\bZ)} \nonumber
\end{equation}
Here we can see that the posterior $p(\bbf,\tilde{\bu},\bw\vert \by, \bX, \bZ)$ assumes all data $\mathbfcal{D}\equiv(\bX, \by)$ is given, and further, its closed form is intractable due to the likelihood 
\begin{equation} 
p(\by\vert\bX,\bZ)=\int p(\by\vert\bbf)p(\bbf\vert\tilde{\bu},\bw,\bX,\bZ)p(\tilde{\bu}\vert\bZ)p(\bw)d\bbf d\tilde{\bu}d\bw \nonumber
\end{equation}
of which integration is intractable due to the presence of the spike-and-slab prior $p(\bw)$. 

In the next section, we discuss a posterior inference framework for the proposed hierarchical model, which tackles the challenges mentioned above.

\subsection{Federated Variational Inference} \label{s:federatedvi}
This section discusses our proposed inference framework. The core idea of our approach is to approximate the original posterior using an estimated surrogate distribution without centralized computation. Consequently, we can use the surrogate distribution in place of the original posterior which causes an issue due to the need to collect all data. We first discuss the surrogate distribution and an optimization problem to be solved for its inference (Section \ref{s:vi}). Then we delve into an FL-based algorithm for the optimization problem, which facilitates the use of local computing power and shares only needed information while preserving privacy (Section \ref{s:federated}).

\subsubsection{Variational inference for sparse LMC}\label{s:vi} 
Our approach is based on VI \cite{Blei2017}, a method for approximate posterior inference that estimates a surrogate distribution to approximate the true posterior distribution. This estimation is often done by solving an optimization problem that minimizes the Kullback-Leibler (KL) divergence between the surrogate and the original posterior. It can be shown that minimizing the KL divergence is indeed equivalent to maximizing a lower bound of the log-likelihood, often referred to as the evidence lower bound (ELBO). By choosing a surrogate distribution with manageable forms, such as factorized Gaussian, the ELBO becomes tractable, whereby a gradient-based optimization algorithm can be used to maximize. Interestingly, the use of VI for the inference of our model built upon the hierarchical structure allows for deriving an ELBO adequate for a federated optimization algorithm.

\textbf{Variational distributions.} We first introduce variational distributions that serve as surrogates to approximate the posterior $p(\bff, \tilde \bu, \bw)$. Due to the presence of the Dirac delta distribution in the spike-and-slab prior $p(\bw)$ in \eqref{eq:ssp}, it is not straightforward to build a proper surrogate that well approximates the posterior distribution for $\bw$. Thus, we reparameterize $\bw$ to make its form more manageable. Specifically, the reparameterization considers $w_{m,l} = \tilde{w}_{m,l}\alpha_l$ for all $m,l$ with which the prior is rewritten as
\begin{align}
p(\tilde{\bw},\balpha)&=\prod_{m=1}^M\prod_{l=1}^Lp(\tilde{w}_{m,l})p(\alpha_{l}) \nonumber\\
&=\prod_{m=1}^M\prod_{l=1}^L\cN(\tilde{w}_{m,l}; 0,\sigma_\bw^2)\pi^{\alpha_{l}}(1-\pi)^{1-\alpha_{l}} \nonumber
\end{align}
for which $\tilde{w}_{m,l}\sim\cN\left(0, \sigma^2_{\bw}\right)$, $\alpha_{l}\sim\pi^{\alpha_{l}}(1-\pi)^{1-\alpha_{l}}$, and $\balpha=[\alpha_l]^\top_{l=1,\cdots,L}$ where $\alpha_{l}\in \{0,1\}$. 

Given the reparameterization, we introduce a surrogate distribution of $p(\bbf, \tilde{\bu},\tilde \bw,\balpha)$ presented as 
\begin{align} \label{eq:surrogate}
q(\bbf, \tilde{\bu},\tilde \bw,\balpha ) =\prod_{m=1}^M p(\bbf_m|\tilde{\bu},\tilde \bw_m, \balpha)q(\tilde{\bu})q(\tilde \bw_m, \balpha) %\label{eq:q(f,u,w)} 
\end{align}
where 
\begin{align} 
&p(\bbf_m|\tilde{\bu},\tilde \bw_m, \balpha )\nonumber\\&\>=\cN\left(\bbf_m; \sum_{l=1}^L(\tilde w_{m,l} \alpha_{l}) \bA_{m,l}\tilde{\bu}_l,  \sum_{l=1}^L(\tilde w_{m,l} \alpha_{l})^2\bK_{m,l}\right),\label{p(f_m|u)} 
\end{align}
\begin{align} 
q(\tilde{\bu}) =\prod_{l=1}^L q(\tilde{\bu}_l) = \prod_{l=1}^L\cN(\tilde{\bu}_l; \bmu_{l}, \bSigma_l),\label{q(u)}
\end{align}
\begin{align} 
&q(\tilde \bw, \balpha) = \prod_{m=1}^{M}\prod_{l=1}^Lq(\tilde w_{m,l}, \alpha_{l}) \nonumber\\ 
&\>=\prod_{m=1}^{M}\prod_{l=1}^L\cN(\tilde w_{m,l}; \alpha_{l}\mu_{w_{m,l}},\nonumber\\
&\quad\quad\quad\quad\quad\quad\quad
\alpha_{l}\sigma_{w_{m,l}}^2+(1-\alpha_{l})\sigma_\bw^2)\times\gamma_{l}^{\alpha_{l}}(1-\gamma_{l})^{1-\alpha_{l}}.\label{eq:q(w,a)}
\end{align}
Equation \eqref{eq:q(w,a)} presents the joint surrogate distribution for $\tilde \bw$ and $\balpha$ with parameters  $\bmu_{\bw_m}=[\mu_{w_{m,l}}]_{l=1,\cdots,L}^\top$, $\bsigma_{\bw_m}=[\sigma_{w_{m,l}}]_{l=1,\cdots,L}^\top$ for $m\in\{1,...,M\}$, and $\bgamma=[\gamma_{l}]_{l=1,\cdots,L}^\top$, implying that $\tilde w_{m,l}$ follows $\cN(\mu_{m,l}, \sigma^2_{w_{m,l}})$ with probability of $\gamma_l$, otherwise follows $\cN(0, \sigma^2_\bw)$. Given \eqref{p(f_m|u)} and \eqref{q(u)}, 
we can further derive  
\begin{align} \label{eq:q(f|w,a)}
&q(\bff_m \vert \tilde \bw_m, \balpha)
= \int p(\bff_m \vert \tilde \bu, \tilde \bw_m, \balpha)q(\tilde \bu)d \tilde \bu \nonumber\\
&\>=\cN\left(\bff_m;\sum_{l=1}^L(\tilde w_{m,l}\alpha_{l}) \bA_{m,l} \bmu_l, \right. \nonumber\\
&\left. \quad\quad\quad\quad\quad\quad\quad\quad \sum_{l=1}^L (\tilde w_{m,l}\alpha_{l})^2\left(\bK_{m,l} + \bA_{m,l} \bSigma_l \bA_{m,l}^\top \right) \right). 
\end{align} 

In fact, approximating the posterior using the variational distribution \eqref{eq:q(w,a)} defined jointly on $\tilde \bw, \balpha$ has substantial benefits compared to assuming independence $q(\tilde \bw, \balpha) = q(\tilde \bw)q(\balpha)$. The joint variational distribution allows for the modeling of the conditional dependencies between $\tilde \bw$ and $\balpha$, which can capture the $2^M$-component multi-modal distributions. In contrast, assuming independence between $\tilde \bw$ and $\balpha$ yields a uni-modal variational distribution that has a limited ability to approximate the true posterior distribution that is multi-modal. Refer to the work \cite{Titsias2011} for more details.

\textbf{Deriving the ELBO.} Now let us derive the ELBO. Given the surrogate distributions \eqref{eq:surrogate}, VI aims to minimize  $\KL\left(q(\bbf,\tilde{\bu},\tilde{\bw},\balpha)\Vert p(\bbf,\tilde{\bu},\tilde{\bw},\balpha\vert\by)\right)$. This is equivalent to maximizing the lower bound of the log-likelihood derived by
\begin{align}
&\log p(\by\vert\bX, \bZ) \nonumber\\
&\>= \log \int p(\by\vert\bff)p(\bff\vert \tilde \bu, \tilde \bw, \balpha,  \bX, \bZ) p(\tilde \bu \vert \bZ) p(\tilde \bw, \balpha) d\bff d\tilde \bu d\bw d\balpha\nonumber\\
&\>\ge \bbE_{q(\bbf,\tilde{\bu},\tilde{\bw},\balpha)}\left[\log\frac{p(\by\vert\bff)p(\bff\vert \tilde \bu, \tilde \bw, \balpha,  \bX, \bZ) p(\tilde \bu \vert \bZ) p(\tilde \bw, \balpha)}{q(\bbf,\tilde{\bu},\tilde{\bw},\balpha)}\right] \nonumber\\
&\>= \cL_{\elbo}(\bTheta) \nonumber
\end{align}
where $\cL_{\elbo}(\bTheta)$ represents the ELBO derived using the Jensen's inequality and $\bTheta = \{\btheta,\{\bpsi_m\}_{m=1}^M\}$ represents the set of parameters to be estimated. More specifically, $\btheta = \{\{\bmu_l\}_{l=1}^L,\{\bSigma_l\}_{l=1}^L,\{\bxi_l\}_{l=1}^L, \bgamma\}$ are the \textit{global parameters} solely associated with global components shared across units and $\bpsi_m = \{\bmu_{\bw_m},\bsigma_{\bw_m},\bsigma_m\}$ are the \textit{personalized parameters} specific to unit $m$ only. The distinction between global and local parameters plays an important role in the development of our FL-based inferential algorithm, which we will discuss shortly. It is easy to reorganize the ELBO as 
\begin{align}
&\cL_{\elbo}(\bTheta) = \bbE_{q(\tilde \bw, \balpha )}\bbE_{q(\bff|\tilde\bw, \balpha)}[\log p(\by|\bbf)]- \nonumber\\
& \qquad\qquad\quad  \KL(q(\tilde{\bw},\balpha)\Vert p(\tilde{\bw},\balpha)) -\KL(q(\tilde{\bu})\Vert p(\tilde{\bu})) 
\label{eq:elbo} 
\end{align} which should 
be maximized in terms of $\bTheta$ for model inference. By maximizing \eqref{eq:elbo}, the first term, the expected log-likelihood of the data under $q(\bbf)$, renders the predictive curve for each unit fits its data better. At the same time, the second and third terms regularize the variational distributions by penalizing their deviations from prior distributions. In particular, the second term encourages $q(\tilde \bw, \balpha)$ to be close to the spike-and-slab prior $p(\tilde \bw, \balpha)$, resulting in sparsity over inferred coefficients. 

In the next section, we discuss an FL-based algorithm that maximizes \eqref{eq:elbo} through collaborative efforts across units, exploiting their local computing power while keeping unit-specific information confidential.

\subsubsection{Federated optimization}\label{s:federated}
Now let us discuss our algorithm to maximize $\cL_{\elbo}(\bTheta)$. The key idea is based on the fact that $\cL_{\elbo}(\bTheta)$ can be expressed as a summation of $M$ terms, where each term is \textit{independent} of the unit-specific information of other units. To see this, first note that the conditional likelihood $\log p(\by|\bbf)$ in the first term of \eqref{eq:elbo} can be factorized over units $\log p(\by|\bbf) = \sum_{m=1}^M\log p(\by_{m}|\bbf_{m})$. This allows for rewriting \eqref{eq:elbo} by $\cL_{\elbo}(\bTheta) = \sum_{m=1}^M\cV_m(\bpsi_m, \btheta; \mathbfcal{D}_m)$ where
\begin{align}\label{eq:v_m}
&\cV_m(\bpsi_m, \btheta; \mathbfcal{D}_m) =\bbE_{q(\tilde \bw_m, \balpha )}\bbE_{q(\bff_m|\tilde \bw_m, \balpha)}[\log p(\by_m|\bbf_m)] \nonumber\\
& \> \quad\quad- \KL(q(\tilde{\bw}_m,\balpha)\Vert p(\tilde{\bw}_m,\balpha))-r_m \sum_{l=1}^L\KL(q(\tilde{\bu}_l)\Vert p(\tilde{\bu}_l))
\end{align} 
with a closed form available (derived in the Appendix), and $r_m = \frac{N_m}{\sum_{m=1}^MN_m}$ is the weight for unit $m$ proportional to $N_m$.
Here it becomes clear that $\cV_m (\bpsi_m, \btheta; \mathbfcal{D}_m)$ does not involve any specific information of the other units $m' \neq m$. Yet, only the common knowledge on $\tilde \bu_l$ summarized in $\btheta$ consists of the shared part across $\cV_1(\bpsi_1, \btheta; \mathbfcal{D}_1)$ to $\cV_M(\bpsi_M, \btheta; \mathbfcal{D}_M)$. 

The decomposability allows us to propose an FL-based algorithm to maximize $\cL_{\elbo}(\bTheta)$, as presented in Algorithm \ref{alg:federated_update}. The key idea of our algorithm is to let each unit locally minimize its objective $\cV_m(\bpsi_m, \btheta; \mathbfcal{D}_m)$, while units regularly share $\btheta$ with the central server to reach a consensus on the common knowledge through iterative communications. Each iteration of this process is called a communication round. The key steps of a communication round are explained below.

\begin{enumerate}
    \item \textit{Broadcast}: The central server broadcasts the global parameter $\btheta^{h-1}$ to the units $m=1,...,M$.
    \item \textit{Local update}: Upon receiving $\btheta^{h-1}$, unit $m$ executes $\texttt{local\_update}(\btheta^{h-1}, \bpsi_m^{h-1};\mathbfcal{D}_m)$ to obtain a set of updated parameters $\{\btheta^{h}, \bpsi_m^{h}\}$. A fixed number of steps in a gradient-based method are executed to minimize \eqref{eq:v_m} for the local updates. The local computing resource of unit $m$ is exploited to calculate gradients $\Delta_{\bpsi_m, \btheta} \cV(\bpsi_m, \btheta; \mathbfcal{D}_m)$.
    \item \textit{Central update}: The central server collects locally updated global parameters $\btheta^{h}_1,...,\btheta^{h}_M$ from all units, and then executes $\texttt{central\_update}(\btheta^{h}_1,...,\btheta^{h}_M)$ to aggregate the collected parameters into $\btheta^h$. A common approach is weighted averaging: $\btheta^h = \sum_{m=1}^M \frac{N_m}{N} \btheta^{h}_m$. This process can be viewed as achieving consensus on shared knowledge.
\end{enumerate}
This process is repeated until an exit condition, such as total communication rounds, is met. The communication between the central server and units only involves sharing $\btheta^h$ that summarizes common knowledge of the latent patterns. Raw datasets $\mathbfcal{D}_m$ and unit-specific information $\bpsi_m$ are never disclosed to the central server. Thus, this enhances the privacy of each unit and reduces potential latency from transmitting large volumes of raw data. Adding to that, the central server merely serves as a parameter aggregator whereby no substantial computing resource is required. Rather, the framework relies on the collaborative and distributed efforts of participating units, utilizing their local computing power for model training. Finally, we note that our iterative algorithm starts with an initial solution found by \texttt{pre\_processing($\cdot$)}. This initialization is also done without data sharing or centralized computation. Empirically, it often provides a good starting point for our algorithm. Please see details in the Appendix.

\begin{algorithm}{
\BlankLine
\caption{Federated optimization for $\cL_{\sf ELBO}(\bTheta)$}
\label{alg:federated_update}
\SetKwInOut{Input}{Input}
\SetKwInOut{Output}{Output}
\SetKwInOut{Return}{Return}
\Input{data ($\mathbfcal{D}$); initial parameters (${\btheta}^{\sf init}, \{{\bpsi}_m^{\sf init}\}_{m=1}^M$);  total communication rounds ($H$)}
\Output{$\btheta^{H}$, $\{\bpsi_m^{H}\}_{m=1}^M$}
\DontPrintSemicolon
\BlankLine
\SetNoFillComment
    ${\btheta}^0, \{{\bpsi}_m^0\}_{m=1}^M \leftarrow $\texttt{pre\_processing(${\btheta}^{\sf init}, \{{\bpsi}_m^{\sf init}\}_{m=1}^M$)}
    
    \For{$h \leftarrow 1$ \KwTo $H$}{
        The central server broadcasts $\btheta^{h-1}$ to all units $m=1,...,M$

        \For{$m \leftarrow 1$ \KwTo $M$}{
            $\btheta_m^h, \bpsi_m^h \leftarrow \texttt{local\_update}(\btheta^{h-1}, \bpsi_m^{h-1};\mathbfcal{D}_m)$ \\
            \tcp*{At the local units}
        }
        All clients send $\btheta_m^h$ to the central server 
        
        $\btheta^h \leftarrow \texttt{central\_update}(\btheta_1^h,...,\btheta_M^h)$ \\
        \tcp*{At the central server}
    }
\Return{$\btheta^H$,$\{\bpsi_m^H\}_{m=1}^M$}
}
\end{algorithm}

Our federated optimization algorithm stands out from standard FL algorithms by dividing model parameters into global and personalized parameters. Only the locally updated global parameters are shared with the central server for aggregation. This enhances the confidentiality of units at the edge since the personalized parameters containing unit-specific information are not disclosed. Also, this approach enables our model to extract global trends shared across units through the estimation of global parameters, and simultaneously, personalizing the model for each unit by capturing unit-specific features present in their data through the estimation of personalized parameters.

Finally, we note that the local update at each unit is scalable, even for large local datasets of size $N_m$. The log-likelihood in the first term of \eqref{eq:v_m} is decomposable over individual data points, expressed as $\log p(\by_m|\bbf_m) = \sum_{n=1}^{N_m}\log p(y_{m,n}|f_{m,n})$. This facilitates the implementation of stochastic optimization, allowing each unit to compute gradient estimates using only a subset of the data. Specifically, a unit can get a noisy estimate of the gradient $\Delta_{\bpsi_m, \btheta} \cV(\bpsi_m, \btheta; \mathbfcal{D}_m)$ from a randomly selected batch from the local dataset $\mathbfcal{D}_m$. The reader interested in the detailed derivation of \eqref{eq:v_m} for stochastic gradient-based methods is referred to the Appendix.

\subsection{An Efficient Learning Approach for New Units} \label{s:new_units}
Based on our framework, we develop an efficient approach for incorporating new units into the system after model estimation is completed. A straightforward method to address this is by retraining an MGP with an updated dataset that includes both existing and new units. However, this approach requires all units to participate in the learning process each time a new unit is added, resulting in significant inefficiency. In contrast, our approach utilizes only the computing power of the new unit for its integration. The key assumption is that if the new unit's data shares the common latent pattern characterized by the existing units, it is sufficient to use the shared parameters estimated from the existing units and only infer the parameters specific to the new unit. Furthermore, this approach benefits from our automatic latent variable selection, enhancing efficiency by using only the selected latent functions instead of all latent functions.

Suppose we have estimated parameters $\hat \bTheta$ and a set of selected latent functions $l \in \cS \subset \{1,...,L\}$ inferred from existing units. Now, consider a new unit $p \notin \{1,...,M\}$. As useful latent functions are already selected, we can build an LMC that also includes the new unit without the spike-and-slab prior.
The ELBO for this model can be written as 
$\cL^p_{\sf ELBO} (\bpsi_p; \hat \bTheta) = \sum_{m \in \{1,...,M,p\}}\bbE_{q(\bff_m)}[\log p(\by_m|\bbf_m)] -\sum_{l\in \cS} \KL(q(\tilde{\bu}_l)\Vert p(\tilde{\bu}_l))$ 
where now we have $q(\bff_m) = \int p(\bff_m|\tilde \bu)q(\tilde \bu) d\tilde \bu$, with $p(\bff_m|\tilde \bu)$ and $q(\tilde{\bu})$ defined similarly as \eqref{eq:p(f|w,g)} and \eqref{q(u)}, respectively, but $\bw$ is treated as a parameter rather than a random variable and $\tilde{\bu}_l$ is used for $l\in S$ instead of all $l\in \{1,...,L\}$. 

If the commonality across the existing and new units is similar to the commonality among the existing units, we can continue using $\hat \bTheta=\{\hat\btheta,\{\hat\bpsi_m\}_{m=1}^M\}$ when learning the new unit $p$. Thus, it would be sufficient to maximize $\cL^p_{\sf ELBO} (\bpsi_p; \hat \bTheta)$ with respect to $\bpsi_p$ only, while keeping $\hat \bTheta$ fixed. Removing the terms in $\cL^p_{\sf ELBO}$ that are independent of $\bpsi_p$, the optimization problem simplifies to
\begin{equation}\label{eq:newlearning}
\max_{\bpsi_p} \bbE_{q(\bff_p)}[\log p(\by_p|\bbf_p)]
\end{equation}
with $\hat \btheta$ being fixed. Notably, the objective function in \eqref{eq:newlearning} does not involve any parameters or data from other units $m \in \{1,...,M\}$. It is solely related to the new unit $p$ and the estimated global parameters $\hat \btheta$ that remain fixed throughout optimization. Therefore, solving \eqref{eq:newlearning} can be locally performed using the computing power of unit $p$, without the involvement of other units.

\subsection{Prediction}
Once model parameters and variational distributions are estimated, each unit can make predictions for a new input. Let $\by_{m,*}$ and $\bff_{m,*}$ represent observations and function values at $N_*$ new inputs $\bX_{m,*} \in \bbR^{N_* \times d}$ for unit $m$, respectively. The predictive distribution can be written as 
\begin{align}
&p(\by_{m,*}\vert \bX_{m,*}, \mathbfcal{D})\nonumber \\
&\> = \int p(\by_{m,*} |\bff_{m,*})p(\bff_{m,*} \vert \tilde \bu, \tilde \bw_m, \balpha)p(\tilde \bu \vert \mathbfcal{D} ) p(\tilde \bw_m, \balpha \vert \mathbfcal{D} ) \nonumber\\
&\> \hspace{0.63\linewidth} \times d\bff_{m,*} d\tilde \bu d\tilde \bw_m d\balpha \nonumber\\
&\> \approx \int p(\by_{m,*}|\bff_{m,*})q(\bff_{m,*} \vert \tilde\bw_m, \balpha)q(\tilde \bw_m, \balpha)d\bff_{m,*}d\tilde \bw_m d\balpha \nonumber\\
&\> = \int\cN\left(\by_{m,*}; \sum_{l=1}^L(\tilde w_{m,l}\alpha_l) \bA^*_{m,l} \hat\bmu_l, \right.\nonumber\\
&\left. \quad\quad \sum_{l=1}^L (\tilde w_{m,l}\alpha_l)^2\left(\bK^*_{m,l} + \bA^*_{m,l} \hat\bSigma_l {\bA^*_{m,l}}^\top\right)+\hat\bsigma_{m}^2 \right) \nonumber \\
&\hspace{0.58\linewidth} q(\tilde \bw_m, \balpha) d\tilde \bw_m d\balpha \label{eq:prediction}
\end{align}
where $\bA^*_{m,l}$ and $\bK^*_{m,l}$ are calculated similarly to $\bA_{m,l}$ and $\bK_{m,l}$ but with $\bX_{m,*}$; and we use the hat notation for the estimated parameters. To address the intractable integral in \eqref{eq:prediction}, we resort to the Monte Carlo method \cite{murray}. This leads to the following approximation 
\begin{align}
    \bbE[\by_m^*] &\approx \frac{1}{S} \sum_{s=1}^{S} \sum_{l=1}^L w_{m,l}^{(s)}\alpha_{l}^{(s)}\bA^{*}_{m,l} \hat\bmu_l, \label{eq:pred_approx_1}\\
    \var [\by_m^*] &\approx \frac{1}{S^2} \sum_{s=1}^{S} \sum_{l=1}^L ( w_{m,l}^{(s)}\alpha_{l}^{(s)})^2\left(\bK^{*}_{m,l} + \bA^*_{m,l} \hat\bSigma_l {\bA^{*\top}_{m,l}}\right)+\hat\bsigma_{m}^2 \label{eq:pred_approx_2}
\end{align}
where $w_{m,l}^{(s)}, \alpha_{l}^{(s)} \overset{\text{i.i.d.}}{\sim} q(\tilde w_{m,l}, \alpha_l)$ in \eqref{eq:q(w,a)} with the estimated parameters $\hat\mu_{w_{m,l}}$,$\hat\sigma^2_{w_{m,l}}$ and $\hat\gamma_{l}$, for $s=1,...,S$. 

%\cN(\hat\mu_{w_{m,l}},\hat\sigma_{w_{m,l}}^2)$ for $s=1,...,S$.

It is crucial to note that calculating \eqref{eq:pred_approx_1} and \eqref{eq:pred_approx_2} is independent of other units $m' \neq m$. These calculations involve only the information specific to unit $m$ and the common information shared across all units. As a result, predictions at new locations for each unit can be derived locally using its own computing resources, without needing to share information between units. This beneficial feature is achieved by substituting the posterior distributions with their corresponding variational distributions in \eqref{eq:prediction}, which are estimated through our FL framework based on the collaborative efforts of all units.

\section{Experiments} \label{s:experiment}
In this section, we evaluate our proposed model using both simulation and real-world data. Our model, denoted by \texttt{FedLMC-SS}, is compared with the benchmark models summarized below. 
\begin{itemize}
\item \texttt{IGP}: A single-output GP regression model for individual outputs without any inter-output information exchange.
\item \texttt{LMC}: An MGP model with LMC modeling under the centralized setting. 
\item \texttt{LMC-SS}: The centralized LMC with spike-and-slab priors on the coefficients. Comparing our model to this model can highlight our model's competitive predictive performance achieved without data sharing while using local computing capability. 
\item \texttt{FedLMC}: A federated MGP model \cite{Chung2023} with the LMC construction. Comparing our model to this model can underscore the use of the spike-and-slab prior for enhanced predictions by selecting only the needed number of latent functions. 
\end{itemize}

We use the Adam optimizer \cite{kingma2014} for all gradient-based parameter updates. In federated models, we use averaging to aggregate locally updated parameters at the central server. Evaluation is based on performance metrics such as prediction accuracy, the number of selected latent functions, and time consumed during model training. In particular, prediction accuracy is assessed using the mean squared error (MSE) of curve prediction. 

We use the Radial Basis Function (RBF) kernel for all MGPs across all experiments:
\begin{equation}
k_l(\bx, \bx'; \bxi_l) = \sigma_{\sf RBF}^2\exp\left(-\frac{\Vert\bx - \bx'\Vert^2} {2l_{\sf{RBF}}^2}\right) \nonumber
\end{equation}
with $\bxi_l := (\sigma_{\sf{RBF}}, l_{\sf{RBF}})$ where $ \sigma_{\sf{RBF}}^2$ and $l_{\sf{RBF}}$ represent the variance and the length-scale hyperparameters, respectively.  We will discuss important setups and results for each experiment in the subsequent sections, yet more details can be found in the Appendix.

\subsection{Simulation} \label{s:regression}
We perform two experiments using simulated data. In Section \ref{s: regression_simulation}, we investigate models' capabilities when units have missing ranges. In Section \ref{s: learning_simulation}, we test our new unit learning approach.

\subsubsection{Federated regression and missing range extrapolation} \label{s: regression_simulation}
\paragraph{Setup} We generate the dataset using a GP with the following covariance function:
\begin{align}
&\exp\left(-\frac{\bx^2+\bx'^2}{10}\right) \nonumber\\
&\times\left(\cos\left(\bx-\bx'\right)+\cos\left(2\left(\bx-\bx'\right)\right)+\cos\left(3\left(\bx-\bx'\right)\right)\right). \nonumber
\end{align}
This covariance function is characterized by six distinct eigenfunctions. That is, the necessary number of latent functions for the shared pattern across generated curves is six. We generate 10 different curves from the GP. Thus, it corresponds to the situation where we have 10 units in a federated scenario. Here each output consists of 100 observations evenly distributed within $[-5,5]$. We add Gaussian noise to each output value with a standard deviation of 0.2. For one unit (output), we remove observations within an interval with length 3 in $[-5, 5]$. Therefore, we can evaluate if models can make accurate extrapolations for the missing range.

For all MGPs, we employ 10 latent functions, each with 20 inducing points evenly distributed within the input space. We run experiments ten times to evaluate average performances and ensure the consistency of the results. 

\begin{figure*}[htb!]
    \centering \includegraphics[width=\textwidth]{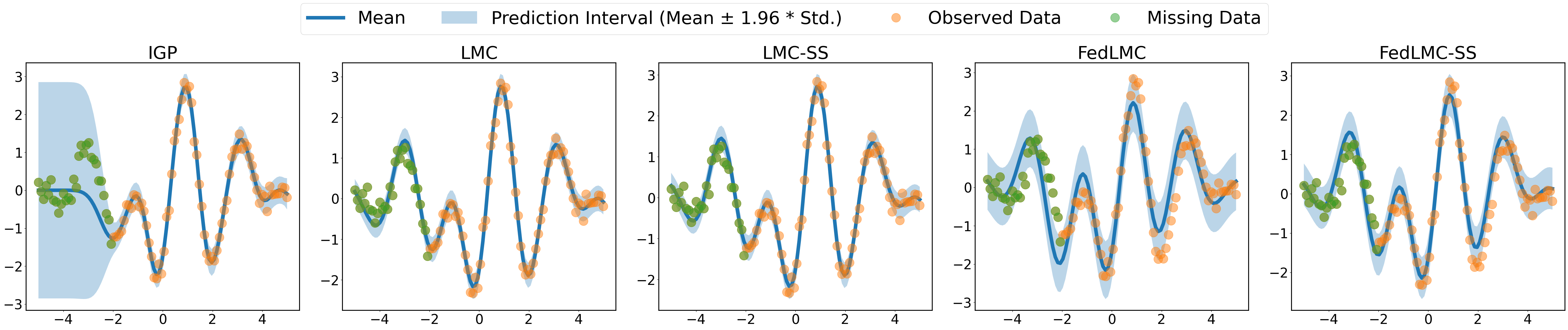}
    \caption{Predictions of the proposed and benchmark models.}
    \label{fig:simulation}
\end{figure*}

\paragraph{Results} Figure \ref{fig:simulation} displays prediction results from one of these runs. Table \ref{tab:simulation} provides the average of MSEs and their standard deviations over repeated experiments.

\begin{table}[htb!]
\centering
\caption{Average MSEs and the number of selected latent functions.}
\resizebox{\columnwidth}{!}{%
\begin{tabular}{@{}ccc@{}}
\toprule
                   & Average MSE ($\pm$Std.) & \# Latent functions ($\pm$Std.)    \\ 
                   \midrule
\texttt{IGP}       & $0.2265\>(\pm0.1486)$   & -                      \\\midrule
\texttt{LMC}       & $0.0536\>(\pm0.0278)$   & $9.1\>(\pm0.9944)$  \\
\texttt{LMC-SS}    & $0.0454\>(\pm0.0170)$   & $6.0\>(\pm0.0000)$  \\
\midrule
\texttt{FedLMC}    & $0.2073\>(\pm0.2249)$   & $10.0\>(\pm0.0000)$ \\
\texttt{FedLMC-SS} & $0.0942\>(\pm0.0741)$   & $6.7\>(\pm1.6364)$  \\ \bottomrule
\end{tabular}
}
\label{tab:simulation}
\end{table}

Several findings can be obtained here. First, in terms of predictive accuracy, both \texttt{LMC-SS} and \texttt{FedLMC-SS} outperform their counterparts \texttt{LMC} and \texttt{FedLMC} without the spike-and-slap priors while selecting a less number of latent functions. The number of latent functions chosen for \texttt{FedLMC-SS} is close to 6, the true number of latent functions used in generating the data. This shows that the spike-and-slab priors allow for automatically selecting only the necessary number of latent functions to describe the shared latent patterns across outputs, resulting in better predictive accuracy by eliminating potential redundancy caused by using too many latent functions. Second, the predictive accuracy of \texttt{FedLMC-SS} is competitive to \texttt{LMC-SS}. This highlights the advantage of our model in federated scenarios. Our model neither needs data sharing across units nor centralized computation while attaining similar predictive accuracy to the centralized model. Third, our model \texttt{FedLMC-SS} significantly outperforms \texttt{IGP}, especially for the range where observations are missing (see Figure \ref{fig:simulation}). This shows that our model successfully leverages information from other units to extrapolate for the missing range, while \texttt{IGP} fails to do so. Note that the extrapolation is achieved without data sharing across units. Finally, Figure \ref{fig:simulation} shows the predictive uncertainty estimated by our model aligns with the distribution of observations.

\subsubsection{Learning for a new unit}\label{s: learning_simulation}
\paragraph{Setup} In this experiment, we examine the learning approach described in Section \ref{s:new_units}. For each run in the previous simulation, we additionally generate 10 new units. For each new unit, we eliminate observations in an interval within $[-5,5]$ with a length of 2. Given a set of selected latent functions and global hyperparameters estimated from the previous 10 units, we learn the new unit's data by solving the optimization problem \eqref{eq:newlearning}. Recall that, during this training process, only local hyperparameters specific to the new units are estimated, while other global components remain fixed. We evaluate the performance of this training strategy by investigating the predictive accuracy of the new units. We examine a method that uses selected latent functions with global parameters estimated by our model \texttt{FedLMC-SS} (denoted as `\texttt{FedLMC-SS}-selected'). This method is compared to two benchmarks: one that uses all latent functions estimated by \texttt{FedLMC} (denoted as `\texttt{FedLMC}-all') and another that uses the latent functions estimated by \texttt{FedLMC} but only employs the same number of latent functions as \texttt{FedLMC-SS}-selected. (denoted as `\texttt{FedLMC}-selected'). For the second case, we determine the most significant latent functions based on the size of the coefficients $\sum_m \Vert w_{m,l} \Vert^2$. 

% Details on settings for the optimization algorithm are deferred to the Appendix. 

\begin{figure}[htb!]
    \centering
    \includegraphics[width=\columnwidth]{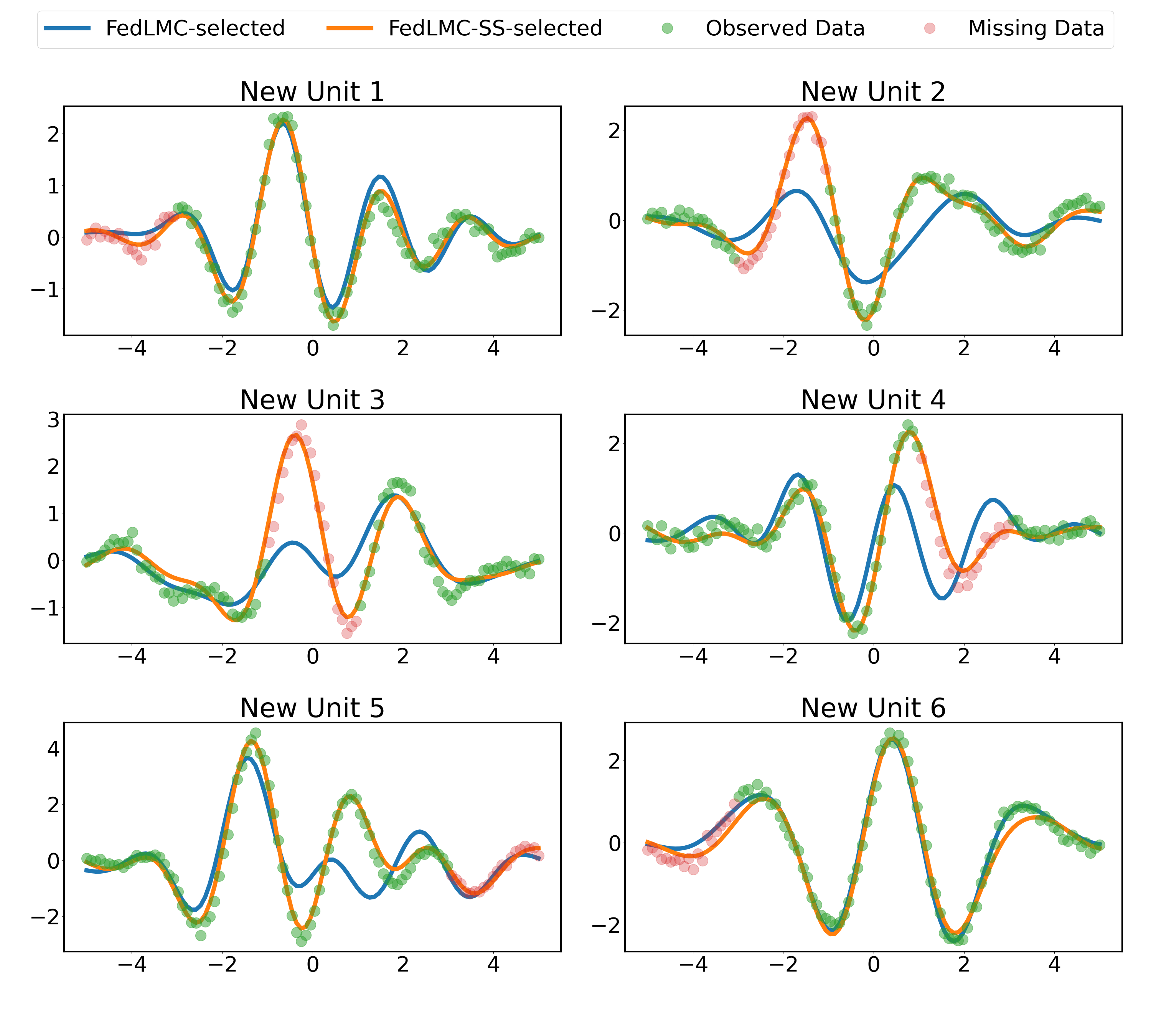}
    \caption{Predictions for the new units using the proposed new unit learning strategy.}
    \label{fig:continual}
\end{figure}

\paragraph{Results} Figure \ref{fig:continual} illustrates the predictions for six new units in an experiment. We have omitted results for the other four units for brevity, as similar trends were observed. Table \ref{tab:continual} summarizes predictive accuracies. The results show that \texttt{FedLMC-SS}-selected performs significantly better than \texttt{FedLMC}-selected. This demonstrates \texttt{FedLMC-SS}'s ability to compress the information of common patterns into fewer latent functions. When using latent functions inferred by \texttt{FedLMC}, predictive accuracy can be increased by using more latent functions, as indicated by the higher accuracy of \texttt{FedLMC}-all compared to \texttt{FedLMC}-selected in Table \ref{tab:continual}. However, this also increases the computational time required to learn the new units. Despite using all latent functions, \texttt{FedLMC}-all is still not competitive to \texttt{FedLMC-SS}-selected. This underperformance is because \texttt{FedLMC} was not able to extract common patterns as effectively as \texttt{FedLMC-SS} even with more latent functions, highlighting the importance of our proposed latent function selection method in achieving more robust latent pattern extraction. Lastly, it is important to note that the model learning and prediction process for the new units did not involve any raw data sharing from units or central computation.

\begin{table}[htb!]
\centering
\caption{Average MSEs ($\times10^{-5}$) and time in seconds per iteration in training.}
    \resizebox{\columnwidth}{!}{
\begin{tabular}{@{}ccc@{}}
\toprule
                & Average MSE ($\pm$Std.) & Time per iteration ($\pm$Std.) \\ \midrule
\texttt{FedLMC}-all         & $0.3101\>(\pm0.2638)$        & $0.0839\>(\pm0.0159)$ \\
\texttt{FedLMC}-selected    & $0.8248\>(\pm0.3446)$        & $0.0644\>(\pm0.0207)$ \\
\texttt{FedLMC-SS}-selected & $0.1521\>(\pm0.1715)$        & $0.0646\>(\pm0.0217)$ \\
\bottomrule
\end{tabular}
    }
\label{tab:continual}
\end{table}

\subsection{Case Studies}
This section discusses experiments using real-world datasets. Our first case study focuses on climate data \cite{Alvarez2008, Parra2017}, specifically air temperature trends from four stations on the southern coast of the UK: Cambermet, Bramblemet, Sotonmet, and Chimet. Due to their proximity, these stations exhibit correlated air temperature trends but with some differences. In our study, each station is treated as a unit. Thus, one can consider an integrative analysis by inferring cross-station correlations inherent in their data. Our objective is to use our model to extract only the necessary latent functions that effectively compress common patterns while ensuring that each station's data remains confidential and is not disclosed to other stations or the central server.

We consider a day in November 2023 when the air temperature is recorded every five minutes. We first train our model \texttt{FedLMC-SS} and the benchmark \texttt{FedLMC}, both with ten latent functions, on the Chimet, Bramblemet, and Sotonmet data. Then, we employ our approach for learning new units to learn the Cambermet data with a missing range using the estimated latent functions. Predictions are made for the missing range, and their accuracies are compared for the cases using the latent functions of \texttt{FedLMC-SS} or \texttt{FedLMC}.

\begin{figure}[htb!]
    \centering
    \includegraphics[width=\columnwidth]{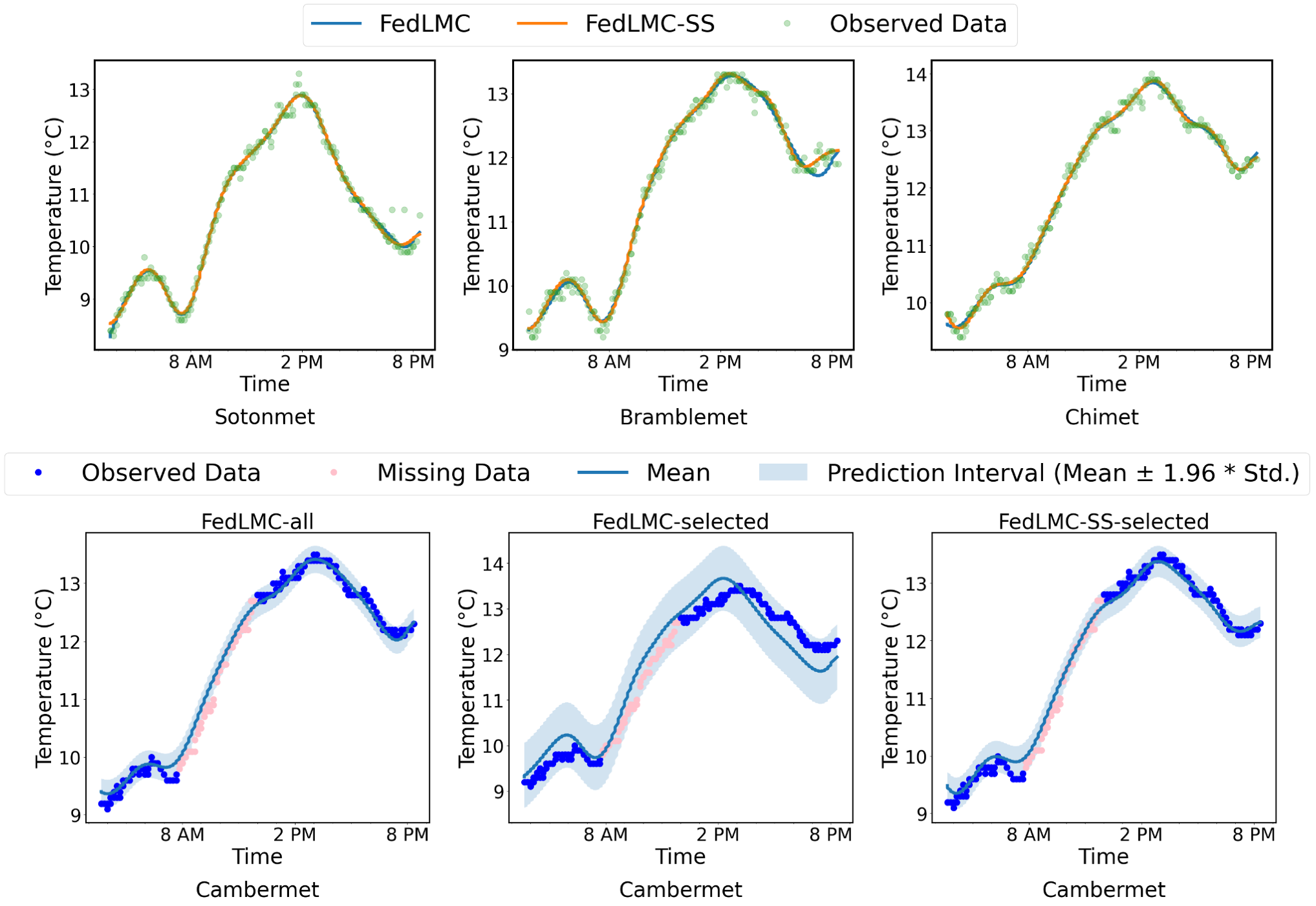}
    \caption{Predictive results for the air temperature data. The first row illustrates predictive curves for Sotonmet, Bramblemet, and Chimet; and the second row presents predictions for Cambermet based on the new unit learning approach.}
    \label{fig:stations}
\end{figure}

Figure \ref{fig:stations} presents the results. The figures in the first row show that both \texttt{FedLMC} and \texttt{FedLMC-SS} effectively fit the data from the three stations. However, the second row illustrates that using the latent functions from our model for the new station (Cambermet) results in significantly better predictions when the same number of latent functions is used. This showcases our method's ability to infer shared patterns in a parsimonious manner. 

\begin{figure}[htb!]
    \centering
    \includegraphics[width=\columnwidth]{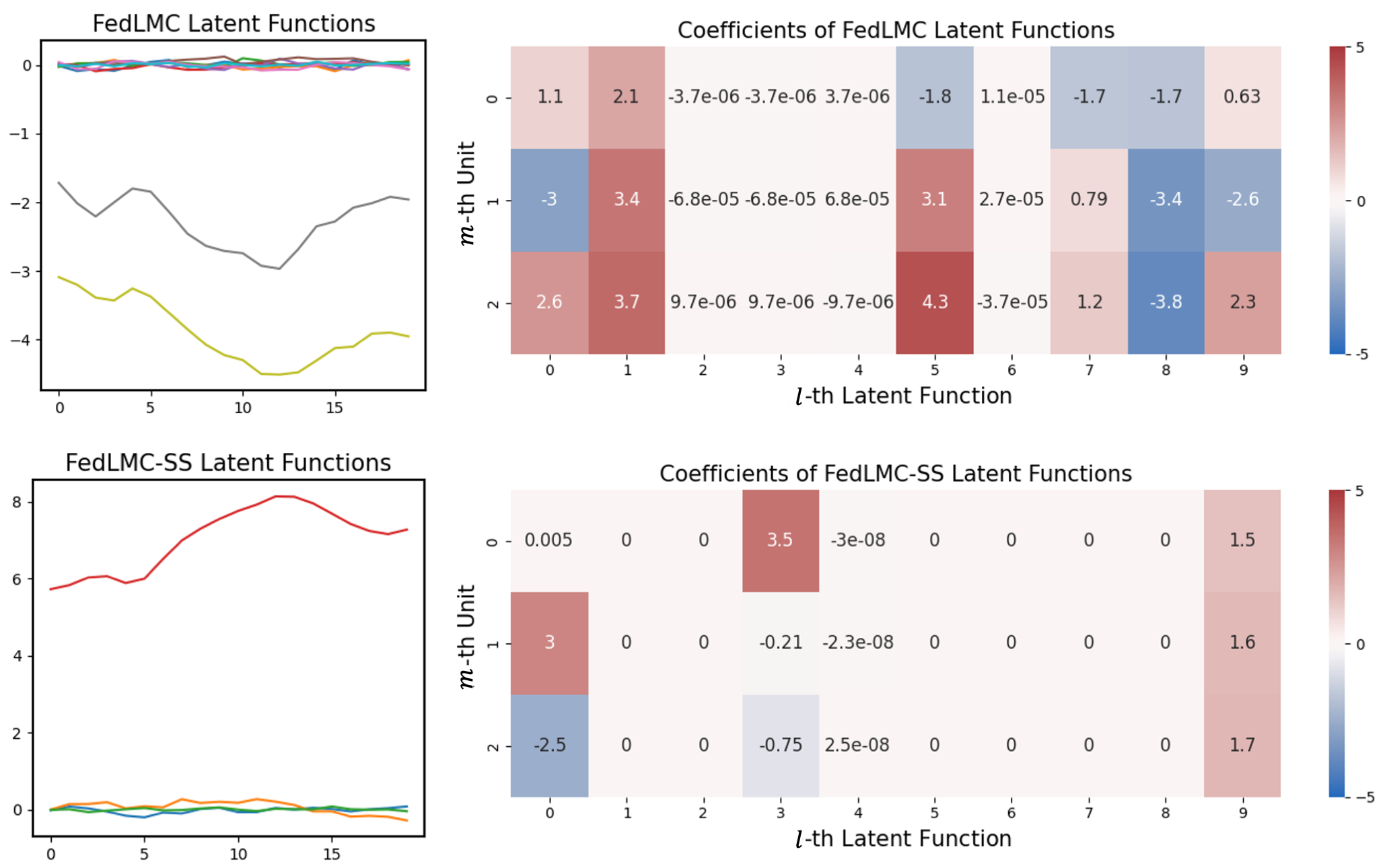}
    \caption{Estimated latent functions and their selection for the air temperature data. The latent function coefficients are regarded as zero if their absolute values are less than $1\times10^{-10}$.}
    \label{fig:latent}
\end{figure}

Additionally, we plot Figure \ref{fig:latent} to show the latent function selection results. The left panel displays the estimated latent function curves, and the right panel presents the heatmaps of the latent function coefficients. In particular, the heatmaps clearly show that \texttt{FedLMC} utilizes more latent functions than \texttt{FedLMC-SS}. This demonstrates that \texttt{FedLMC-SS} more efficiently extracts shared patterns across units. Here it is important to note that discerning each station's data using the estimated latent patterns is challenging. This highlights a key advantage of our approach in enhancing confidentiality, as only the latent patterns that are not useful in inferring local data are shared across units.

Another real-world example focuses on degradation modeling for Li-ion battery cell capacities. In quality engineering, estimating the remaining useful life (RUL) of a system plays a significant role in preventive maintenance. Li-ion batteries have a finite lifetime as their capacity fades over their use. As such, predicting future degradation trends is essential for estimating their RUL to establish an adequate maintenance or replacement plan. MGPs can be utilized as a RUL prediction tool for batteries in use, by leveraging the degradation trends of other batteries with known full lifecycle degradation data. However, in real-world applications, battery degradation data is often collected by different entities (e.g., companies or laboratories) or units (e.g., electric vehicles) that are unwilling to share their data. Additionally, each unit with a battery, such as an electric vehicle, can have local computing capabilities. In this scenario, our proposed method is a promising prediction approach that enhances the privacy of each unit while harnessing its local computing power, as well as, enhancing predictions by effectively learning shared latent curves using sparse priors.

We use the CALCE battery dataset \cite{Diao2020}. It contains capacity fade data for a collection of Li-ion battery cells, each with a nominal capacity of 350 mAh. This dataset includes degradation trends for 23 cells from a production batch, acquired through qualification testing. From these, we randomly select $M=20$ cells. Among these 20 cells, one cell is randomly chosen as the target unit, with its degradation observations available up to the 125th cycle out of a total of 250 cycles. The degradation data for the other 19 cells is available for the full 250 cycles. Furthermore, we examine our new unit learning scheme using the remaining three cells, whose observations are also only available up to the 125th cycle. We repeat this experiment ten times. We build the LMC models with five independent latent functions.

\begin{figure*}[htb!]
    \centering
    \includegraphics[width=\textwidth]{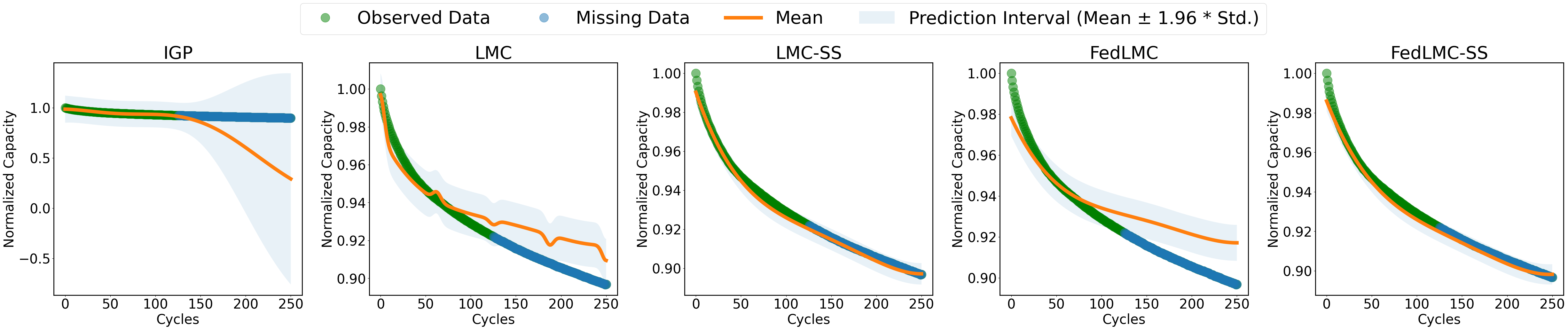}
    \caption{Predictions for the capacity degradation of a battery cell.}
    \label{fig:battery}
\end{figure*}

The results of this case study are presented in Figure \ref{fig:battery} and Table \ref{tab:battery}. One can observe that our \texttt{FedLMC-SS} significantly outperforms \texttt{IGP} while quite competitive to \texttt{LMC-SS}. This sheds light on the benefit of our approach in a federated setting. Specifically, units need not share their degradation data with the central server or other units when leveraging others' degradation information to predict a future degradation curve. In contrast, the independent modeling approach \texttt{IGP} that satisfies a restriction on data sharing fails to make a reasonable prediction. Furthermore, as shown in Table \ref{tab:battery_continual}, the new unit learning approach demonstrates clear benefits for this application.

\begin{table}[htb!]
\centering
\caption{Average MSEs ($\times10^{-5}$) and the number of selected latent functions in the battery degradation case study.}
\resizebox{\columnwidth}{!}{%
\begin{tabular}{@{}ccc@{}}
\toprule
                    & Average MSE ($\pm$Std.) & \# Latent functions ($\pm$Std.)            \\ 
                    \midrule
\texttt{IGP}       & 10302.7028$\>(\pm$506.8795)              & -                                          \\
\midrule
\texttt{LMC}       & 15.2907$\>(\pm$24.0035)                  & $1.9\>(\pm0.8755)$                      \\
\texttt{LMC-SS}    & 0.3682$\>(\pm$0.3596)                    & $2.1\>(\pm0.5676)$                      \\
\midrule
\texttt{FedLMC}    & 22.1664$\>(\pm$22.6758)                  & $1.2\>(\pm0.4216)$                      \\
\texttt{FedLMC-SS} & 0.8535$\>(\pm$0.9157)                    & $2.0\>(\pm0.0000)$ \\ 
\bottomrule
\end{tabular}
}
\label{tab:battery}
\end{table}

% \begin{figure}[htb!]
%     \centering
%     \includegraphics[width=0.75\columnwidth]{figures/battery_continual.png}
%     \caption{Predictions for the new battery cells using the proposed learning strategy}
%     \label{fig:battery_continual}
% \end{figure}

Interestingly, placing a spike-and-slab prior significantly improves predictions over those without the prior, although both methods tend to use one or two latent functions to model degradation curves. We hypothesize that this improvement is because the spike-and-slab prior introduces prior knowledge on the selection of latent functions. Table \ref{tab:battery} suggests that only a few latent functions would be sufficient to model the degradation curves, despite the availability of five potential latent functions. The sparse prior implicitly biases our model towards solutions with fewer latent functions. On the other hand, methods without the prior have to explore the solution space expanded by five latent variables without such guidance. As a result, the model easily gets trapped in poor local optima.
\begin{table}[htb!]
\centering
\caption{Average MSEs ($\times10^{-5}$) and time in seconds ($\times 10^{-3}$) per iteration in training in battery degradation case study.}
\resizebox{\columnwidth}{!}{
\begin{tabular}{@{}ccc@{}}
\toprule
                            & Average MSE ($\pm$Std.)
 & Time per iteration ($\pm$Std.)                \\ 
                            \midrule
\texttt{FedLMC}-selected    & 13.6876$\>(\pm$8.0891)                   & 5.2622$\>(\pm$0.6421)                \\
\texttt{FedLMC-SS}-selected & 0.6092$\>(\pm$0.5191)                    & 6.4215$\>(\pm$0.0394) \\
\bottomrule
\end{tabular}
    }
\label{tab:battery_continual}
\end{table}

\section{Conclusion} \label{s:discussion_conclusion}
Existing MGPs based on LMC suffer key challenges, including the difficulty of setting the correct number of latent functions and the reliance on a centralized training framework. This paper introduces a hierarchical modeling approach that addresses these limitations simultaneously. Our approach represents each unit's data as a linear combination of a common set of independent latent processes shared across units, with coefficients assigned spike-and-slab priors. These priors enable the automatic determination of the number of independent latent processes. Building on VI, we propose an FL-based inference method that utilizes each unit's local computing power while keeping its data reside at the edge. Real-world applications on air temperature trend modeling and quality engineering highlight the advantages of our model compared to standard LMC models that do not select latent functions or rely on centralized model learning.

\bibliographystyle{IEEEtran}
\bibliography{Literature}

\end{document}